\documentclass[11pt]{article}
\usepackage{geometry}
\usepackage{coling2020}
\usepackage{times}
\usepackage{url}
\usepackage{latexsym}
\usepackage{microtype}

\usepackage[hidelinks]{hyperref}

\colingfinalcopy 

\title{BRUMS at SemEval-2020 Task 12 : Transformer based Multilingual Offensive Language Identification in Social Media}

\author{Tharindu Ranasinghe$^1$, Hansi Hettiarachchi$^2$ \\

  $^1$Research Group in Computational Linguistics, University of Wolverhampton, UK \\
  
  $^2$School of Computing and Digital Technology, Birmingham City University, UK \\
   {\tt t.d.ranasinghehettiarachchige@wlv.ac.uk } \\
   {\tt hansi.hettiarachchi@mail.bcu.ac.uk } } 

\date{}

\begin{document}
\maketitle
\begin{abstract}
 In this paper, we describe the team \textit{BRUMS} entry to OffensEval 2: Multilingual Offensive Language Identification in Social Media in SemEval-2020. The OffensEval organizers provided participants with annotated datasets containing posts from social media in Arabic, Danish, English, Greek and Turkish. We present a multilingual deep learning model to identify offensive language in social media. Overall, the approach achieves acceptable evaluation scores, while maintaining flexibility between languages.
\end{abstract}

\section{Introduction}

\blfootnote{
    %
    
    \hspace{-0.65cm}  
    This work is licensed under a Creative Commons 
    Attribution 4.0 International Licence.
    Licence details:
    \url{http://creativecommons.org/licenses/by/4.0/}.
    
}

Social media has become a normal medium of communication for people these days as it provides the convenience of sending messages fast
from a variety of devices. Unfortunately, social networks also provide the means for distributing abusive and aggressive content. Given the amount of information generated every day on social media, it is not possible for humans to identify and remove such messages manually, instead it is necessary to employ automatic methods. As offensive language becomes pervasive in social media, scholars and companies have been working on developing systems capable of identifying offensive posts, which can be set aside for human moderation or permanently deleted \cite{risch-krestel-2018-delete}.

Along with these studies, a few shared tasks have been organized on detecting offense in social media, such as HatEval \cite{basile2019semeval}, HASOC \cite{10.1145/3368567.3368584}, TRAC \cite{kumar-etal-2018-benchmarking} and OffensEval \cite{zampieri-etal-2019-semeval} co-located with SemEval 2019. In semeval 2020, OffensEval returns for the second time with multilingual offensive language identification in social media. OffensEval 2 focuses on several languages Arabic, Danish, English, Greek and Turkish motivating participants to submit multilingual offensive language identification systems.   

This paper revisits the problem of offensive language identification describing our submission to the SemEval-2020 Task 1 \cite{zampieri-etal-2020-semeval}. The remainder of this paper is structured as follows: Section \ref{sec:relwork} describes the related work done in the field of aggression detection, Section \ref{sec:task} has a description of the dataset. Section \ref{sec:meth} describes the system that was submitted, split into a description of how the data was processed and the architectures of the classifiers that were used. Section \ref{sec:res} presents an analysis of the results of our evaluation of the different architectures, as well as of the final submission. Finally, Section \ref{sec:conc} offers some final remarks and a conclusion.

\section{Relevant Work}
\label{sec:relwork}
The majority of text classification approaches in early research work were supported by traditional or feature-based supervised learning methods. Following this tradition, Malmasi and Zampieri \shortcite{malmasi2017detecting} proposed a linear Support Vector Machine (SVM) classifier for hate speech detection in social media. The best accuracy of this approach was obtained by using character n-grams as model features. Another research suggested stacked ensemble system using logistic regression and random forest-based classifiers \cite{montani2018tuwienkbs}. For initial layer of classifiers, they used character n-grams, token n-grams, important tokens and word2vec embeddings. The Outputs of initial layer classifiers were used as the inputs of next classifier layer which produces the final result. This approach obtained the best results in GermEval 2018 \cite{wiegand2018overview} by outperforming the deep neural network (DNN) models such as convolutional neural networks (CNN) \cite{von2018spmmmp} and combination of bi-directional Long-Short term memory and CNN (BiLSTM-CNN) \cite{wiedemann2018transfer}.

However, most of the recent research conducted in this area prove that DNN-based methods can outperform traditional supervised learning methods \cite{zampieri-etal-2019-predicting,modha2018filtering}. But, unlike the traditional methods, DNNs require sufficient amount of training data to adjust their weights properly without overfitting. Focusing on this issue, Modha et al. \shortcite{modha2018filtering} found that LSTM with higher dropout ($\approx 0.5$) can handle overfitting problem. In order to increase the model performance using a high number of training instances in a simple manner, the available data sets related to aggression and cyberbullying detection can be merged \cite{fortuna2018merging}. But, when focused on a particular area, it cannot guarantee the availability of multiple data sets. Another research suggested data augmentation and pseudo labelling approaches to increase the amount of training data \cite{aroyehun2018aggression}. As the data augmentation strategy, they proposed to translate each instance into an intermediate language and back to the original language. For pseudo labelling approach, they trained some initial models using available training data and picked the best model to label new data. A LSTM model trained using the augmented data set received the first place in Trolling, Aggression and Cyberbullying (TRAC-2018) shared task designed for Facebook English data set. In the same shared task, for the platform unnamed English data set, third place could be obtained by a CNN-LSTM model trained using the combination of augmented and pseudo labelled data sets. 

In addition to approaching data set increasing methods, transfer learning techniques also used by previous research to overcome the data limitation issue. Wiedemann et al. \shortcite{wiedemann2018transfer} showed that transfer learning can be applied using some labelled data set with relevant labels or unlabelled data set with unsupervised user clusters or topics. Further, as mentioned in SemEval-2019 OffensEval leader-board, the top 5 solutions for offensive language identification are BERT-based and they proved the effectiveness in transfer learning as well as Transformers \cite{zampieri-etal-2019-semeval}. The top solution of this shared task fined tuned a BERT model using the training data and found that it outperforms a LSTM model with higher dropout \cite{liu2019nuli}. 

In summary, majority of recent research in this area were supported by DNN models and among them, CNN, LSTM and BERT-based models were found to be most effective and commonly used. When focused on the input data representation, there was a tendency to use fastText embeddings to handle the high availability of spelling mistakes and new words in social media \cite{aroyehun2018aggression,modha2018filtering,wiedemann2018transfer,zampieri-etal-2019-predicting}. Following this tendency, in this research, we further experimented the effectiveness of DNN models including some recently published transformers for offensive language identification.

\section{Dataset}
\label{sec:task}
OffensEval 2020, the SemEval 2020 Offensive Language Detection Shared Task, does not provide a new manually labeled training set for the English language \cite{zampieri-etal-2020-semeval}. The competitors were recommended to use the Offensive Language Identification Dataset (OLID) which was released for the the SemEval 2019 Offensive Language Detection Shared Task \cite{zampieri-etal-2019-semeval}. The OLID is a hierarchical dataset to identify the type and the target of offensive texts in social media. The dataset is collected on Twitter and publicly available. There are 14,100 tweets in total, in which 13,240 are in the training set, and 860 are in the test set. For each tweet,there are three levels of labels:(A) Offensive/Not-Offensive, (B) Targeted-Insult/Untargeted, (C) Indi-vidual/Group/Other. For English subtask organisers provided an additional dataset similar to OLID \cite{zampieri-etal-2019-semeval}, it still has three levels, but this time only confidence scores, generated by different models, are provided rather than human annotated labels. In addition, the data in level A is separated from levels B and C. In level A, there are 9,089,140 tweets,in levels B and C, there are different 188,973 tweets.  

For non-English languages the organisers released Multilingual Offensive Language Identification Dataset (MOLID). It contains four languages: Arabic \cite{mubarak2020arabic}, Danish \cite{sigurbergsson-derczynski-2020-offensive}, Greek \cite{Zeses-etal-2020-offensive}, and Turkish \cite{coltekin-2020-corpus}. Table \ref{tab:data} contains information about number of rows that the training, dev and test datasets had in each language \cite{zampieri-etal-2020-semeval}. 

\begin{table}
\centering
\begin{tabular}{ |c|c|c|c| } 
\hline
Language & training & dev & test \\ 
\hline
Arabic & 7000 & 1000 & 2000 \\ 
\hline
Danish & 2961 & - & 330 \\ 
\hline
Greek & 8744 & - & 1545 \\ 
\hline
Turkish & 31757 & - & 3529 \\ 
\hline
\end{tabular}
\caption{Number of rows in Training set(training), Development set(dev) and Test set(test) in each language. "-" denotes that no data was available.}
\label{tab:data}
\end{table}

\section{Methodology}
\label{sec:meth}
In this section we present the methodology employed for the shared task. All the languages followed a common methodology which had two steps: Data Preprocessing and Machine Learning. 

\subsection{Data Preprocessing}
As mentioned previously, the data preprocessing for this task was kept fairly minimal to make it portable for all the languages. More specifically, we perform only four specialised tasks for this data, followed by tokenisation. The tasks include removing usernames, removing \textit{URL}s, converting all the emojis to text and converting all tokens to lower case. 

First, we completely remove all usernames from the tweets, without inserting a placeholder. In the SemEval-2020 dataset, mention of a certain user is represented by \textit{@USER}. We removed all strings containing \textit{@USER} using a regular expression. The reasoning behind this step is mainly to remove noisy text. Then we remove all \textit{URL} mentions in the tweets. In the SemEval-2020 dataset, mention of a certain URL is represented by \textit{URL}. We removed all strings containing \textit{URL} using a regular expression. The reasoning behind this step too is mainly to remove noisy text. Emojis play an important role in showing aggression in social media \cite{hettiarachchi-ranasinghe-2019-emoji}. Since we can't guarantee that the pretrained embedding models we use will have embeddings for emjois, we converted all the emojis to text using a third party python library named \textit{Emoji} \footnote{https://github.com/carpedm20/emoji/}. We only conducted this step for English, since the \textit{Emoji} library does not support other languages.

For Arabic language, we used a separate preprocessing step. Twitter users writing in Arabic can either write in Modern Standard Arabic (MSA) or in their particular dialects. If the tweet is written in MSA, it may or may not include diacritics whereas dialectical Arabic does not include any. Offensive tweets can be formulated in any of these different versions of the Arabic transcript. Thus, in order to avoid false classifications due to a non-defining feature, we used regular expressions to delete diacritics.

Final prepossessing step is only applied to the architectures that used character embeddings. The fastText pretrained character embedding models that we used only contain lower-cased letters. Therefore,  we convert the text to lower case letters. However, for the Transformer architecture, we did this preprocessing step only if the model that we used is uncased.  

\subsection{Machine Learning}
In order to determine the most suitable neural network architecture for the task, we experimented with three different neural architectures: Convolutional Neural Network (CNN) \cite{kim-2014-convolutional}, Recurrent Neural Network (RNN) \cite{10.1007/978-3-030-00018-9_15} and Transformer \cite{devlin2018bert}. 

\subsubsection{Convolutional Neural Network (CNN)}
\label{subsec:cnn}
CNN is a class of deep, feed-forward artificial neural networks that uses a variation of multilayer perceptrons \cite{kim-2014-convolutional}. CNNs are generally used in computer vision, however they have recently been applied to various NLP tasks and the results were promising \cite{kim-2014-convolutional,hughes2017medical}. In a CNN result of each convolution will fire when a special pattern is detected. By varying the size of the kernels and concatenating their outputs, we can detect patterns of multiples sizes (2, 3, or 5 adjacent words) \cite{kim-2014-convolutional}. Patterns could be expressions like “fuck it”, “stupid cunt” and CNNs can identify them in the sentence regardless of their position. Since offense is mostly a word pattern, we assumed that CNNs would be a good architecture to detect offensive sentences.

For this architecture we used fasttext character embeddings \cite{bojanowski2016enriching}. After the embedding layer we used four convolution layers with 1,2,3 and kernal sizes followed by a max pooling layer,  \cite{10.1007/978-3-642-15825-4_10}. Finally the outputs of the max pooling layers were concatenated and passed through a linear layer. For all the languages in the task, we used fasttext character embeddings released by facebook \cite{grave2018learning} \footnote{\url{https://fasttext.cc/docs/en/crawl-vectors.html}}.

\subsubsection{Recurrent Neural Network (RNN)}
\label{subsec:rnn}
RNNs are designed to make use of sequential data, when the current step has some kind of relation with the previous steps. \cite{10.1007/978-3-030-00018-9_15}. This makes them ideal for applications with a time component (audio, time-series data) and natural language processing \cite{Zeses-etal-2020-offensive}. RNNs perform very well for applications where sequential information is clearly important, because the meaning could be misinterpreted if sequential information is not used. Since sequential information is important to identify offensive sentences, we used the RNN architecture for this task. 
For this architecture too we used fasttext character embeddings \cite{bojanowski2016enriching} trained on English Wikipedia text. After the embedding layer we used a RNN layer. We experimented with LSTM \cite{10.1162/neco.1997.9.8.1735}, bi-directional LSTM \cite{10.5555/1986079.1986220}, GRU \cite{69e088c8129341ac89810907fe6b1bfe} and bi-directional GRU \cite{69e088c8129341ac89810907fe6b1bfe}. The output of the RNN layer was followed a linear layer finally. For this architecture also we used fasttext character embeddings released by facebook.

\subsubsection{Transformers}
\label{subsec:transformers}
With the introduction of BERT \cite{devlin2018bert} 
transformer architectures have shown a massive success in wide range of NLP tasks. Transformer architectures have been trained on general tasks like language modelling and then fine-tuned for classification tasks. \cite{10.1007/978-3-030-32381-3_16,ranasinghe2019brums}. 

Transformer models take an input of a sequence and outputs the representation of the sequence. The sequence has one or two segments that the first token of the sequence is always [CLS] which contains
the special classification embedding and another
special token [SEP] is used for separating segments.
For text classification tasks, Transformer models take the final hidden state \textbf{h} of the first token [CLS] as the representation of the whole sequence \cite{10.1007/978-3-030-32381-3_16}. A simple
softmax classifier is added to the top of the transformer model  to predict the probability of label c as shown in Equation \ref{equ:softmax} where W is the task-specific parameter matrix
. 
\begin{equation}
\label{equ:softmax}
p(c|\textbf{h}) = softmax(W\textbf{h}) 
\end{equation}

We fine-tuned all the parameters from transformer models as well as W jointly by maximising the log-probability of the correct label. For English language, there exists many pretrained transformer models released by Hugging Face\footnote{\url{https://huggingface.co}}. Therefore, we experimented several transformer architectures like BERT \cite{devlin2018bert}, XLNet \cite{yang2019xlnet}, XLM \cite{conneau2019unsupervised}, RoBERTa \cite{liu2019roberta} and DistilBERT \cite{sanh2019distilbert}. We used the HuggingFace's implementation of the transformer models \cite{Wolf2019HuggingFacesTS} and the pre-trained models available in the HuggingFace's model repository\footnote{\url{https://huggingface.co/models}}. However, for non-English languages, Google only provides a single BERT multilinugal model that supports 104 languages\footnote{\url{https://github.com/google-research/bert/blob/master/multilingual.md}}. DistilBERT also has a multilingual model that supports the same 104 languages that multilingual BERT supports. We used these two models for all the non-English languages. To experiment more, we also experimented several community built transformer models. 

For Arabic, other than the multilingual BERT model and multilingual DistilBERT we used AraBERT \cite{antoun2020arabert} \footnote{\url{https://github.com/aub-mind/arabert}}. AraBERT is an Arabic pretrained lanaguage model based on Google's BERT architechture. The model was trained on ~70M sentences or ~23GB of Arabic text with ~3B words. The training corpora are a collection of publicly available large scale raw Arabic text (Arabic Wikidumps, The 1.5B words Arabic Corpus \cite{ElKhair201615BW}, The OSIAN Corpus \cite{zeroual-etal-2019-osian}, Assafir news articles, and 4 other manually crawled news websites (Al-Akhbar, Annahar, AL-Ahram, AL-Wafd). There are two version off the model AraBERTv0.1 and AraBERTv1, with the difference being that AraBERTv1 uses pre-segmented text where prefixes and suffixes were splitted using the Farasa Segmenter \cite{abdelali-etal-2016-farasa}. We used AraBERTv1 since it performed better from the two models in most of the downstream NLP tasks \cite{antoun2020arabert}.

For Danish language, we used Danish-BERT \footnote{\url{https://github.com/botxo/nordic_bert}}, a Danish pretrained language model based on BERT architecture. The Danish corpus used to train the Danish-BERT was compiled by combining multiple sources: All Danish language text from Common Crawl, The Danish Wikipedia, Custom scraped data from the two biggest Danish debate forums (dindebat.dk and hestenettet.dk) and Danish OpenSubtitles.

We used GreekBERT\footnote{\url{https://github.com/nlpaueb/greek-bert}} for Greek Language. This model too has been trained on combining multiple text sources: Greek Wikidumps, Greek part of European Parliament Proceedings Parallel Corpus and Greek language text from OSCAR \cite{ortizsuarez:hal-02148693}, a cleansed version of Common Crawl.Common Crawl. Unlike the transformer models we used for other languages GreekBERT is uncased. Therefore, we had to convert all tokens to lowercase before feeding them to the model. 

For Turkish we used  BERTurk\footnote{\url{https://github.com/stefan-it/turkish-bert}} which is also based on the BERT architecture. The current version of the model was trained on a filtered and sentence segmented version of the Turkish OSCAR corpus \cite{ortizsuarez:hal-02148693}, a recent Wikipedia dump and various OPUS corpora \cite{tiedemann-2012-parallel}.

\section{Results}
\label{sec:res}
In this section we present the evaluation results that were obtained during testing for each language. We also provide a brief look at the final submission results of the shared task. In all the languages multilingual DistilBERT does not perform better than multilingual BERT model we used. Therefore, we have omitted it from the results. 

\subsection{Arabic}
For Arabic language, we trained our system on the training set and evaluated the system on development data that the organisers provided. Table \ref{table:arabic} shows the results we obtained for the development set. As shown Transformer models outperform traditional word embedding models. Also it should be noted that AraBERT slightly outperformed multilingual BERT. On the test set our system achieved 0.788 macro F1 score and ranked 41\textsuperscript{st} out of 53 teams. 

\begin{table*}[htb]
\centering
\begin{tabular}{l|ccc|ccc|ccc|c}

\hline
                                     & \multicolumn{3}{c|}{\textbf{Not Offensive}} & \multicolumn{3}{c|}{\textbf{Offensive}}             & \multicolumn{3}{c|}{\textbf{Weighted Average}}      & \textbf{}         \\ \hline
\multicolumn{1}{l|}{\textbf{Model}} & \textbf{P}   & \textbf{R}   & \textbf{F1}   & \textbf{P} & \textbf{R} & \textbf{F1}               & \textbf{P} & \textbf{R} & \textbf{F1}               & \textbf{F1 Macro} \\ \hline
\textit{CNN} & 0.80         & 0.82         & 0.81          & 0.71       & 0.66       & 0.69 & 0.82       & 0.82       & 0.78 & 0.70    \\
\textit{RNN-BILSTM} & 0.76         & 0.78         & 0.77          & 0.67       & 0.62       & 0.65  & 0.78       & 0.78       & 0.74 & 0.66              \\
\textit{BERT-multilingual-cased}  & 0.82         & 0.84         & 0.83          & 0.74       & 0.69       & 0.72  & 0.86       & 0.87       & 0.86 & 0.77
\\
\textit{AraBERT}  & 0.84  & 0.86  & 0.85          & 0.76  & 0.71         & 0.74  & 0.88       & 0.89       & 0.88 & \textbf{0.79}

          \\ \hline
\end{tabular}
\caption[Results for Arabic subtask ]{Results for English Task For each model, Precision (P), Recall (R), and F1 are reported on all classes, and weighted averages. Macro-F1 is also listed (best in bold).}
\label{table:arabic}
\end{table*}

\subsection{Danish}
Since the organisers didn't provide a separate development set for Danish language, we separated 20\% of the training data and treated it as the development set to evaluate the models. Table \ref{table:danish} shows the results. In Danish language also transformer models outperformed traditional word embedding models. However, Danish-BERT model could not outperform the BERT Multilingual model. We suspect that this is mainly due to the fact that Danish-BERT model we used was uncased. For the test data our best system scored 0.656 Macro F1 score ranking 32\textsuperscript{nd} out of 39 teams.  

\begin{table*}[htb]
\centering
\begin{tabular}{l|ccc|ccc|ccc|c}

\hline
                                     & \multicolumn{3}{c|}{\textbf{Not Offensive}} & \multicolumn{3}{c|}{\textbf{Offensive}}             & \multicolumn{3}{c|}{\textbf{Weighted Average}}      & \textbf{}         \\ \hline
\multicolumn{1}{l|}{\textbf{Model}} & \textbf{P}   & \textbf{R}   & \textbf{F1}   & \textbf{P} & \textbf{R} & \textbf{F1}               & \textbf{P} & \textbf{R} & \textbf{F1}               & \textbf{F1 Macro} \\ \hline
\textit{CNN} & 0.91  & 0.92   & 0.93  & 0.71   & 0.67     & 0.67 & 0.85       & 0.85       & 0.85 & 0.72 \\
\textit{RNN-BILSTM} & 0.90         & 0.91         & 0.92          & 0.70       & 0.66       & 0.66 & 0.84       & 0.84       & 0.84 & 0.71  \\
\textit{BERT-multilingual-cased} & 0.95         & 0.96         & 0.97          & 0.75       & 0.75       & 0.75 & 0.93       & 0.92       & 0.93 & \textbf{0.80} \\
\textit{DANISH-BERT} & 0.92     & 0.93         & 0.93          & 0.71       & 0.71       & 0.71 & 0.89       & 0.89       & 0.89 & 0.76 
          \\ \hline
\end{tabular}
\caption[Results for Danish subtask ]{Results for English Task For each model, Precision (P), Recall (R), and F1 are reported on all classes, and weighted averages. Macro-F1 is also listed (best in bold).}
\label{table:danish}
\end{table*}

\subsection{English}
For English language also we separated an evaluation set which was 20\% from the training set and evaluated our models on that. Since there were many pretrained transformer models for English, we were able to experiment with various transformer architectures. Table \ref{table:english} shows the results for the evaluation set. \textit{XLNET large cased} transformer model outperformed all the other architectures. Our best system scored 0.90056 Macro F1 score for the test set and ranked 62\textsuperscript{nd} out of 86 teams. 
\begin{table*}[htb]
\centering
\begin{tabular}{l|ccc|ccc|ccc|c}

\hline
                                     & \multicolumn{3}{c|}{\textbf{Not Offensive}} & \multicolumn{3}{c|}{\textbf{Offensive}}             & \multicolumn{3}{c|}{\textbf{Weighted Average}}      & \textbf{}         \\ \hline
\multicolumn{1}{l|}{\textbf{Model}} & \textbf{P}   & \textbf{R}   & \textbf{F1}   & \textbf{P} & \textbf{R} & \textbf{F1}               & \textbf{P} & \textbf{R} & \textbf{F1}               & \textbf{F1 Macro} \\ \hline
\textit{CNN}  & 0.87         & 0.88         & 0.87          & 0.81       & 0.77       & 0.77 & 0.81       & 0.83       & 0.82 & 0.79     \\
\textit{RNN-BILSTM} & 0.86         & 0.87         & 0.86          & 0.80       & 0.76       & 0.76 & 0.80       & 0.81       & 0.81 & 0.77              \\
\textit{BERT-large-cased}   & 0.91         & 0.92         & 0.91          & 0.85       & 0.81       & 0.81  & 0.85       & 0.85       & 0.86 & 0.82     \\
\textit{XLNet-large-cased}    & 0.92         & 0.93         & 0.92          & 0.84       & 0.83       & 0.83  & 0.87       & 0.87       & 0.88 & \textbf{0.85} 
          \\ \hline
\end{tabular}
\caption[Results for English Task ]{Results for English Task For each model, Precision (P), Recall (R), and F1 are reported on all classes, and weighted averages. Macro-F1 is also listed (best in bold).}
\label{table:english}
\end{table*}

\subsection{Greek}
For Greek we prepared a separate evaluation set as we did with Danish and English since the organisers did not provide a development set. In Greek, GreekBERT performed best outperforming multilingual BERT model slightly. Table \ref{table:greek} shows the results for the evaluation set. For the test set it had 0.814 Macro F1 score and ranked 15\textsuperscript{th} out of 37 teams.
\begin{table*}[htb]
\centering
\begin{tabular}{l|ccc|ccc|ccc|c}

\hline
                                     & \multicolumn{3}{c|}{\textbf{Not Offensive}} & \multicolumn{3}{c|}{\textbf{Offensive}}             & \multicolumn{3}{c|}{\textbf{Weighted Average}}      & \textbf{}         \\ \hline
\multicolumn{1}{l|}{\textbf{Model}} & \textbf{P}   & \textbf{R}   & \textbf{F1}   & \textbf{P} & \textbf{R} & \textbf{F1}               & \textbf{P} & \textbf{R} & \textbf{F1}               & \textbf{F1 Macro} \\ \hline
\textit{CNN} & 0.81         & 0.82         & 0.83          & 0.74       & 0.75       & 0.74 & 0.86       & 0.85       & 0.84 & 0.80     \\
\textit{RNN-BILSTM}   & 0.80         & 0.80         & 0.81          & 0.73       & 0.72       & 0.73 & 0.77       & 0.77       & 0.75 & 0.71              \\
\textit{BERT-multilingual-cased}   & 0.87         & 0.87         & 0.88          & 0.79       & 0.79       & 0.79 & 0.83       & 0.83       & 0.82 & 0.77     \\
\textit{GREEKBERT}              & 0.89         & 0.89         & 0.90          & 0.81       & 0.81       & 0.81 & 0.85       & 0.85       & 0.85 & \textbf{0.81} 
          \\ \hline
\end{tabular}
\caption[Results for Greek subtask ]{Results for Greek subtask For each model, Precision (P), Recall (R), and F1 are reported on all classes, and weighted averages. Macro-F1 is also listed (best in bold).}
\label{table:greek}
\end{table*}

\subsection{Turkish}
For Turkish also we prepared a separate evaluation set as we did with Danish and English. For Turkish, the transformer model with BERTURK performed best surpassing BERT multilingual as shown in table \ref{table:Turkish}. Our best system had 0.7858 Macro F1 score and ranked 7\textsuperscript{th} out of 46 teams.

\begin{table*}[htb]
\centering
\begin{tabular}{l|ccc|ccc|ccc|c}

\hline
                                     & \multicolumn{3}{c|}{\textbf{Not Offensive}} & \multicolumn{3}{c|}{\textbf{Offensive}}             & \multicolumn{3}{c|}{\textbf{Weighted Average}}      & \textbf{}         \\ \hline
\multicolumn{1}{l|}{\textbf{Model}} & \textbf{P}   & \textbf{R}   & \textbf{F1}   & \textbf{P} & \textbf{R} & \textbf{F1}               & \textbf{P} & \textbf{R} & \textbf{F1}               & \textbf{F1 Macro} \\ \hline
\textit{CNN}                  & 0.80         & 0.81         & 0.79          & 0.70       & 0.59       & 0.69 & 0.75       & 0.76       & 0.75 & 0.67    \\
\textit{RNN-BILSTM}                     & 0.79         & 0.80         & 0.78          & 0.69       & 0.58       & 0.68 & 0.74       & 0.75       & 0.74 & 0.66              \\
\textit{BERT-multilingual-cased} & 0.83         & 0.84         & 0.82          & 0.73       & 0.62       & 0.72 & 0.78       & 0.79       & 0.78 & 0.70    \\
\textit{BERTURK}              & 0.92         & 0.93         & 0.91          & 0.82       & 0.82       & 0.81 & 0.87       & 0.88       & 0.87 & \textbf{0.79} 
          \\ \hline
\end{tabular}
\caption[Results for Turkish subtask]{Results for Turkish subtask For each model, Precision (P), Recall (R), and F1 are reported on all classes, and weighted averages. Macro-F1 is also listed (best in bold).}
\label{table:Turkish}
\end{table*}

\section{Conclusion}
\label{sec:conc}
In this paper, we have presented the team \textit{BRUMS} system for identifying offensive language in social media posts in Arabic, Danish,English, Greek and Turkish. The system uses minimal preprocessing, and relies on word and context embeddings. We experimented with different deep neural network architectures in order to determine the most suitable for this task. Our implementation has been made available on Github.\footnote{\url{https://github.com/tharindudr/offenseval_2020}}. According to our evaluation, and the results provided by the task organisers, it is clear that fine tuning transformer architectures score highest overall. Also we discovered that pretrained monolingual transformer models perform better than the \textit{bert-multilingual} model. In future, we are hoping to experiment more with transformer architectures in different languages.

\bibliographystyle{coling}
\bibliography{semeval2020}

\begin{thebibliography}{}

\bibitem[\protect\citename{Abdelali \bgroup et al.\egroup
  }2016]{abdelali-etal-2016-farasa}
Ahmed Abdelali, Kareem Darwish, Nadir Durrani, and Hamdy Mubarak.
\newblock 2016.
\newblock {F}arasa: A fast and furious segmenter for {A}rabic.
\newblock In {\em Proceedings of the 2016 Conference of the North {A}merican
  Chapter of the Association for Computational Linguistics: Demonstrations},
  pages 11--16, San Diego, California, June. Association for Computational
  Linguistics.

\bibitem[\protect\citename{Antoun \bgroup et al.\egroup
  }2020]{antoun2020arabert}
Wissam Antoun, Fady Baly, and Hazem Hajj.
\newblock 2020.
\newblock Arabert: Transformer-based model for arabic language understanding.

\bibitem[\protect\citename{Aroyehun and Gelbukh}2018]{aroyehun2018aggression}
Segun~Taofeek Aroyehun and Alexander Gelbukh.
\newblock 2018.
\newblock Aggression detection in social media: Using deep neural networks,
  data augmentation, and pseudo labeling.
\newblock In {\em Proceedings of the First Workshop on Trolling, Aggression and
  Cyberbullying (TRAC-2018)}, pages 90--97.

\bibitem[\protect\citename{Basile \bgroup et al.\egroup
  }2019]{basile2019semeval}
Valerio Basile, Cristina Bosco, Elisabetta Fersini, Debora Nozza, Viviana
  Patti, Francisco Manuel~Rangel Pardo, Paolo Rosso, and Manuela Sanguinetti.
\newblock 2019.
\newblock Semeval-2019 task 5: Multilingual detection of hate speech against
  immigrants and women in twitter.
\newblock In {\em Proceedings of the 13th International Workshop on Semantic
  Evaluation}, pages 54--63.

\bibitem[\protect\citename{Bojanowski \bgroup et al.\egroup
  }2016]{bojanowski2016enriching}
Piotr Bojanowski, Edouard Grave, Armand Joulin, and Tomas Mikolov.
\newblock 2016.
\newblock Enriching word vectors with subword information.
\newblock {\em arXiv preprint arXiv:1607.04606}.

\bibitem[\protect\citename{Chung \bgroup et al.\egroup
  }2014]{69e088c8129341ac89810907fe6b1bfe}
Junyoung Chung, Caglar Gulcehre, Kyunghyun Cho, and Yoshua Bengio.
\newblock 2014.
\newblock Empirical evaluation of gated recurrent neural networks on sequence
  modeling.
\newblock In {\em NIPS 2014 Workshop on Deep Learning, December 2014}.

\bibitem[\protect\citename{{\c{C}}{\"o}ltekin}2020]{coltekin-2020-corpus}
{\c{C}}a{\u{g}}r{\i} {\c{C}}{\"o}ltekin.
\newblock 2020.
\newblock A corpus of {T}urkish offensive language on social media.
\newblock In {\em Proceedings of The 12th Language Resources and Evaluation
  Conference}, pages 6174--6184, Marseille, France, May. European Language
  Resources Association.

\bibitem[\protect\citename{Conneau \bgroup et al.\egroup
  }2019]{conneau2019unsupervised}
Alexis Conneau, Kartikay Khandelwal, Naman Goyal, Vishrav Chaudhary, Guillaume
  Wenzek, Francisco Guzm{\'a}n, Edouard Grave, Myle Ott, Luke Zettlemoyer, and
  Veselin Stoyanov.
\newblock 2019.
\newblock Unsupervised cross-lingual representation learning at scale.
\newblock {\em arXiv preprint arXiv:1911.02116}.

\bibitem[\protect\citename{Cui \bgroup et al.\egroup
  }2018]{10.1007/978-3-030-00018-9_15}
Jianjing Cui, Jun Long, Erxue Min, Qiang Liu, and Qian Li.
\newblock 2018.
\newblock Comparative study of cnn and rnn for deep learning based intrusion
  detection system.
\newblock In Xingming Sun, Zhaoqing Pan, and Elisa Bertino, editors, {\em Cloud
  Computing and Security}, pages 159--170, Cham. Springer International
  Publishing.

\bibitem[\protect\citename{Devlin \bgroup et al.\egroup }2018]{devlin2018bert}
Jacob Devlin, Ming-Wei Chang, Kenton Lee, and Kristina Toutanova.
\newblock 2018.
\newblock Bert: Pre-training of deep bidirectional transformers for language
  understanding.
\newblock {\em arXiv preprint arXiv:1810.04805}.

\bibitem[\protect\citename{El-Khair}2016]{ElKhair201615BW}
Ibrahim~Abu El-Khair.
\newblock 2016.
\newblock 1.5 billion words arabic corpus.
\newblock {\em ArXiv}, abs/1611.04033.

\bibitem[\protect\citename{Fortuna \bgroup et al.\egroup
  }2018]{fortuna2018merging}
Paula Fortuna, Jos{\'e} Ferreira, Luiz Pires, Guilherme Routar, and S{\'e}rgio
  Nunes.
\newblock 2018.
\newblock Merging datasets for aggressive text identification.
\newblock In {\em Proceedings of the First Workshop on Trolling, Aggression and
  Cyberbullying (TRAC-2018)}, pages 128--139.

\bibitem[\protect\citename{Grave \bgroup et al.\egroup
  }2018]{grave2018learning}
Edouard Grave, Piotr Bojanowski, Prakhar Gupta, Armand Joulin, and Tomas
  Mikolov.
\newblock 2018.
\newblock Learning word vectors for 157 languages.
\newblock In {\em Proceedings of the International Conference on Language
  Resources and Evaluation (LREC 2018)}.

\bibitem[\protect\citename{Graves \bgroup et al.\egroup
  }2005]{10.5555/1986079.1986220}
Alex Graves, Santiago Fern\'{a}ndez, and J\"{u}rgen Schmidhuber.
\newblock 2005.
\newblock Bidirectional lstm networks for improved phoneme classification and
  recognition.
\newblock In {\em Proceedings of the 15th International Conference on
  Artificial Neural Networks: Formal Models and Their Applications - Volume
  Part II}, ICANN’05, page 799–804, Berlin, Heidelberg. Springer-Verlag.

\bibitem[\protect\citename{Hettiarachchi and
  Ranasinghe}2019]{hettiarachchi-ranasinghe-2019-emoji}
Hansi Hettiarachchi and Tharindu Ranasinghe.
\newblock 2019.
\newblock Emoji powered capsule network to detect type and target of offensive
  posts in social media.
\newblock In {\em Proceedings of the International Conference on Recent
  Advances in Natural Language Processing (RANLP 2019)}, pages 474--480, Varna,
  Bulgaria, September. INCOMA Ltd.

\bibitem[\protect\citename{Hochreiter and
  Schmidhuber}1997]{10.1162/neco.1997.9.8.1735}
Sepp Hochreiter and J\"{u}rgen Schmidhuber.
\newblock 1997.
\newblock Long short-term memory.
\newblock {\em Neural Comput.}, 9(8):1735–1780, November.

\bibitem[\protect\citename{Hughes \bgroup et al.\egroup
  }2017]{hughes2017medical}
M~Hughes, I~Li, S~Kotoulas, and T~Suzumura.
\newblock 2017.
\newblock Medical text classification using convolutional neural networks.
\newblock {\em Studies in health technology and informatics}, 235:246.

\bibitem[\protect\citename{Kim}2014]{kim-2014-convolutional}
Yoon Kim.
\newblock 2014.
\newblock Convolutional neural networks for sentence classification.
\newblock In {\em Proceedings of the 2014 Conference on Empirical Methods in
  Natural Language Processing ({EMNLP})}, pages 1746--1751, Doha, Qatar,
  October. Association for Computational Linguistics.

\bibitem[\protect\citename{Kumar \bgroup et al.\egroup
  }2018]{kumar-etal-2018-benchmarking}
Ritesh Kumar, Atul~Kr. Ojha, Shervin Malmasi, and Marcos Zampieri.
\newblock 2018.
\newblock Benchmarking aggression identification in social media.
\newblock In {\em Proceedings of the First Workshop on Trolling, Aggression and
  Cyberbullying ({TRAC}-2018)}, pages 1--11, Santa Fe, New Mexico, USA, August.
  Association for Computational Linguistics.

\bibitem[\protect\citename{Liu \bgroup et al.\egroup }2019a]{liu2019nuli}
Ping Liu, Wen Li, and Liang Zou.
\newblock 2019a.
\newblock Nuli at semeval-2019 task 6: transfer learning for offensive language
  detection using bidirectional transformers.
\newblock In {\em Proceedings of the 13th International Workshop on Semantic
  Evaluation}, pages 87--91.

\bibitem[\protect\citename{Liu \bgroup et al.\egroup }2019b]{liu2019roberta}
Yinhan Liu, Myle Ott, Naman Goyal, Jingfei Du, Mandar Joshi, Danqi Chen, Omer
  Levy, Mike Lewis, Luke Zettlemoyer, and Veselin Stoyanov.
\newblock 2019b.
\newblock Roberta: A robustly optimized bert pretraining approach.
\newblock {\em arXiv preprint arXiv:1907.11692}.

\bibitem[\protect\citename{Malmasi and Zampieri}2017]{malmasi2017detecting}
Shervin Malmasi and Marcos Zampieri.
\newblock 2017.
\newblock {Detecting Hate Speech in Social Media}.
\newblock In {\em Proceedings of the International Conference Recent Advances
  in Natural Language Processing (RANLP)}, pages 467--472.

\bibitem[\protect\citename{Mandl \bgroup et al.\egroup
  }2019]{10.1145/3368567.3368584}
Thomas Mandl, Sandip Modha, Prasenjit Majumder, Daksh Patel, Mohana Dave,
  Chintak Mandlia, and Aditya Patel.
\newblock 2019.
\newblock Overview of the hasoc track at fire 2019: Hate speech and offensive
  content identification in indo-european languages.
\newblock In {\em Proceedings of the 11th Forum for Information Retrieval
  Evaluation}, FIRE ’19, page 14–17, New York, NY, USA. Association for
  Computing Machinery.

\bibitem[\protect\citename{Modha \bgroup et al.\egroup
  }2018]{modha2018filtering}
Sandip Modha, Prasenjit Majumder, and Thomas Mandl.
\newblock 2018.
\newblock Filtering aggression from the multilingual social media feed.
\newblock In {\em Proceedings of the First Workshop on Trolling, Aggression and
  Cyberbullying (TRAC-2018)}, pages 199--207.

\bibitem[\protect\citename{Montani}2018]{montani2018tuwienkbs}
Joaqu{\i}n~Padilla Montani.
\newblock 2018.
\newblock Tuwienkbs at germeval 2018: German abusive tweet detection.
\newblock In {\em 14th Conference on Natural Language Processing KONVENS},
  volume 2018, page~45.

\bibitem[\protect\citename{Mubarak \bgroup et al.\egroup
  }2020]{mubarak2020arabic}
Hamdy Mubarak, Ammar Rashed, Kareem Darwish, Younes Samih, and Ahmed Abdelali.
\newblock 2020.
\newblock Arabic offensive language on twitter: Analysis and experiments.

\bibitem[\protect\citename{Ortiz~Su{\'a}rez \bgroup et al.\egroup
  }2019]{ortizsuarez:hal-02148693}
Pedro~Javier Ortiz~Su{\'a}rez, Beno{\^i}t Sagot, and Laurent Romary.
\newblock 2019.
\newblock {Asynchronous Pipeline for Processing Huge Corpora on Medium to Low
  Resource Infrastructures}.
\newblock In {\em {7th Workshop on the Challenges in the Management of Large
  Corpora (CMLC-7)}}, Cardiff, United Kingdom, July.

\bibitem[\protect\citename{Pitenis \bgroup et al.\egroup
  }2020]{Zeses-etal-2020-offensive}
Zeses Pitenis, Marcos Zampieri, and Tharindu Ranasinghe.
\newblock 2020.
\newblock Offensive language identification in greek.
\newblock In {\em Proceedings of the Twelfth International Conference on
  Language Resources and Evaluation ({LREC} 2020)}, Marseille, France, May.
  European Language Resources Association (ELRA).

\bibitem[\protect\citename{Ranasinghe \bgroup et al.\egroup
  }2019]{ranasinghe2019brums}
Tharindu Ranasinghe, Marcos Zampieri, and Hansi Hettiarachchi.
\newblock 2019.
\newblock Brums at hasoc 2019: Deep learning models for multilingual hate
  speech and offensive language identification.
\newblock {\em In Proceedings of the 11th annual meeting of the Forum for
  Information Retrieval Evaluation (December 2019)}.

\bibitem[\protect\citename{Risch and Krestel}2018]{risch-krestel-2018-delete}
Julian Risch and Ralf Krestel.
\newblock 2018.
\newblock {Delete or not Delete? Semi-Automatic Comment Moderation for the
  Newsroom}.
\newblock In {\em Proceedings of the First Workshop on Trolling, Aggression and
  Cyberbullying (TRAC-2018)}, pages 166--176.

\bibitem[\protect\citename{Sanh \bgroup et al.\egroup
  }2019]{sanh2019distilbert}
Victor Sanh, Lysandre Debut, Julien Chaumond, and Thomas Wolf.
\newblock 2019.
\newblock Distilbert, a distilled version of bert: smaller, faster, cheaper and
  lighter.
\newblock {\em arXiv preprint arXiv:1910.01108}.

\bibitem[\protect\citename{Scherer \bgroup et al.\egroup
  }2010]{10.1007/978-3-642-15825-4_10}
Dominik Scherer, Andreas M{\"u}ller, and Sven Behnke.
\newblock 2010.
\newblock Evaluation of pooling operations in convolutional architectures for
  object recognition.
\newblock In Konstantinos Diamantaras, Wlodek Duch, and Lazaros~S. Iliadis,
  editors, {\em Artificial Neural Networks -- ICANN 2010}, pages 92--101,
  Berlin, Heidelberg. Springer Berlin Heidelberg.

\bibitem[\protect\citename{Sigurbergsson and
  Derczynski}2020]{sigurbergsson-derczynski-2020-offensive}
Gudbjartur~Ingi Sigurbergsson and Leon Derczynski.
\newblock 2020.
\newblock Offensive language and hate speech detection for {D}anish.
\newblock In {\em Proceedings of The 12th Language Resources and Evaluation
  Conference}, pages 3498--3508, Marseille, France, May. European Language
  Resources Association.

\bibitem[\protect\citename{Sun \bgroup et al.\egroup
  }2019]{10.1007/978-3-030-32381-3_16}
Chi Sun, Xipeng Qiu, Yige Xu, and Xuanjing Huang.
\newblock 2019.
\newblock How to fine-tune bert for text classification?
\newblock In Maosong Sun, Xuanjing Huang, Heng Ji, Zhiyuan Liu, and Yang Liu,
  editors, {\em Chinese Computational Linguistics}, pages 194--206, Cham.
  Springer International Publishing.

\bibitem[\protect\citename{Tiedemann}2012]{tiedemann-2012-parallel}
J{\"o}rg Tiedemann.
\newblock 2012.
\newblock Parallel data, tools and interfaces in {OPUS}.
\newblock In {\em Proceedings of the Eighth International Conference on
  Language Resources and Evaluation ({LREC}'12)}, pages 2214--2218, Istanbul,
  Turkey, May. European Language Resources Association (ELRA).

\bibitem[\protect\citename{von Gr{\"u}nigen \bgroup et al.\egroup
  }2018]{von2018spmmmp}
Dirk von Gr{\"u}nigen, Fernando Benites, Pius von D{\"a}niken, Mark Cieliebak,
  Ralf Grubenmann, and AG~SpinningBytes.
\newblock 2018.
\newblock spmmmp at germeval 2018 shared task: Classification of offensive
  content in tweets using convolutional neural networks and gated recurrent
  units.
\newblock In {\em 14th Conference on Natural Language Processing KONVENS 2018},
  page 130.

\bibitem[\protect\citename{Wiedemann \bgroup et al.\egroup
  }2018]{wiedemann2018transfer}
Gregor Wiedemann, Eugen Ruppert, Raghav Jindal, and Chris Biemann.
\newblock 2018.
\newblock Transfer learning from lda to bilstm-cnn for offensive language
  detection in twitter.
\newblock {\em arXiv preprint arXiv:1811.02906}.

\bibitem[\protect\citename{Wiegand \bgroup et al.\egroup
  }2018]{wiegand2018overview}
Michael Wiegand, Melanie Siegel, and Josef Ruppenhofer.
\newblock 2018.
\newblock {Overview of the GermEval 2018 Shared Task on the Identification of
  Offensive Language}.
\newblock In {\em Proceedings of GermEval}.

\bibitem[\protect\citename{Wolf \bgroup et al.\egroup
  }2019]{Wolf2019HuggingFacesTS}
Thomas Wolf, Lysandre Debut, Victor Sanh, Julien Chaumond, Clement Delangue,
  Anthony Moi, Pierric Cistac, Tim Rault, R'emi Louf, Morgan Funtowicz, and
  Jamie Brew.
\newblock 2019.
\newblock Huggingface's transformers: State-of-the-art natural language
  processing.
\newblock {\em ArXiv}, abs/1910.03771.

\bibitem[\protect\citename{Yang \bgroup et al.\egroup }2019]{yang2019xlnet}
Zhilin Yang, Zihang Dai, Yiming Yang, Jaime Carbonell, Russ~R Salakhutdinov,
  and Quoc~V Le.
\newblock 2019.
\newblock Xlnet: Generalized autoregressive pretraining for language
  understanding.
\newblock In {\em Advances in neural information processing systems}, pages
  5754--5764.

\bibitem[\protect\citename{Zampieri \bgroup et al.\egroup
  }2019a]{zampieri-etal-2019-predicting}
Marcos Zampieri, Shervin Malmasi, Preslav Nakov, Sara Rosenthal, Noura Farra,
  and Ritesh Kumar.
\newblock 2019a.
\newblock Predicting the type and target of offensive posts in social media.
\newblock In {\em Proceedings of the 2019 Conference of the North {A}merican
  Chapter of the Association for Computational Linguistics: Human Language
  Technologies, Volume 1 (Long and Short Papers)}, pages 1415--1420,
  Minneapolis, Minnesota, June. Association for Computational Linguistics.

\bibitem[\protect\citename{Zampieri \bgroup et al.\egroup
  }2019b]{zampieri-etal-2019-semeval}
Marcos Zampieri, Shervin Malmasi, Preslav Nakov, Sara Rosenthal, Noura Farra,
  and Ritesh Kumar.
\newblock 2019b.
\newblock {S}em{E}val-2019 task 6: Identifying and categorizing offensive
  language in social media ({O}ffens{E}val).
\newblock In {\em Proceedings of the 13th International Workshop on Semantic
  Evaluation}, pages 75--86, Minneapolis, Minnesota, USA, June. Association for
  Computational Linguistics.

\bibitem[\protect\citename{Zampieri \bgroup et al.\egroup
  }2020]{zampieri-etal-2020-semeval}
Marcos Zampieri, Preslav Nakov, Sara Rosenthal, Pepa Atanasova, Georgi
  Karadzhov, Hamdy Mubarak, Leon Derczynski, Zeses Pitenis, and
  \c{C}a\u{g}r{\i} \c{C}\"{o}ltekin.
\newblock 2020.
\newblock {SemEval-2020 Task 12: Multilingual Offensive Language Identification
  in Social Media (OffensEval 2020)}.
\newblock In {\em Proceedings of SemEval}.

\bibitem[\protect\citename{Zeroual \bgroup et al.\egroup
  }2019]{zeroual-etal-2019-osian}
Imad Zeroual, Dirk Goldhahn, Thomas Eckart, and Abdelhak Lakhouaja.
\newblock 2019.
\newblock {OSIAN}: Open source international {A}rabic news corpus - preparation
  and integration into the {CLARIN}-infrastructure.
\newblock In {\em Proceedings of the Fourth Arabic Natural Language Processing
  Workshop}, pages 175--182, Florence, Italy, August. Association for
  Computational Linguistics.

\end{thebibliography}

\end{document}